\documentclass[american,final,a4paper,12pt]{article}  

\usepackage[margin=1in]{geometry}

\usepackage{amsmath,amsfonts,bm}

\newcommand{\net}{\mathcal{G}}

\newcommand{\id}{\mathds{1}}

\newcommand{\act}{a}

\newcommand{\Norm}{\mathcal{N}}

\newcommand{\Prob}{\mathbb{P}}

\def\eqref#1{equation~\ref{#1}}
\def\1{\bm{1}}

\def\eps{{\epsilon}}

\DeclareMathAlphabet{\mathsfit}{\encodingdefault}{\sfdefault}{m}{sl}
\SetMathAlphabet{\mathsfit}{bold}{\encodingdefault}{\sfdefault}{bx}{n}

\newcommand{\E}{\mathbb{E}}

\usepackage[american]{babel}
\usepackage[latin1]{inputenc}
\usepackage{etex}
\usepackage[T1]{fontenc}
\usepackage{marvosym}
\usepackage{relsize}
\usepackage[ruled,vlined]{algorithm2e}

\usepackage{eso-pic} % used by \AddToShipoutPicture
\usepackage{fancyhdr}

\usepackage{mathptmx,charter,courier}
\usepackage[scaled]{helvet}

\usepackage{hyperref}
\hypersetup{
	colorlinks,
	citecolor={blue!70!black},
	linkcolor={blue!70!black},
	urlcolor={blue!70!black}}

\usepackage{url}
\usepackage{dsfont}
\usepackage{graphicx}
\usepackage{caption}
\usepackage{subcaption}
\usepackage{amsthm}
\usepackage{natbib}
\usepackage{setspace}
\usepackage{booktabs}
\usepackage{multirow}
\usepackage{soul}
\usepackage{rotating}

\theoremstyle{definition}
\newtheorem{definition}{Definition}

\newcommand{\exs}[2]{{\mathbb E_{#1}}\left[ #2 \right]}

\newcommand\numberthis{\addtocounter{equation}{1}\tag{\theequation}}

\medskipamount

\usepackage[shortlabels]{enumitem}
\setlist{noitemsep, nolistsep}

\begin{document}

% ====================================================================================
%
\title{Fake News in Social Networks\thanks{We thank Olimpia Carradori for excellent research assistance and Martina Dosser for help with programming the experiment. For comments and suggestions, we thank Immanuel Lampe, Christopher Summerfield, and participants of the European Meeting of the Economic Science Association in Bologna, the Annual Conference of the Swiss Society of Economics and Statistics in Fribourg, and the Thurgau Experimental Economics Meeting in Kreuzlingen. IRB approval was obtained from the University of St.\ Gallen. Financial support from GFF Project Funding from the University of St.\ Gallen is gratefully acknowledged.}
} 
%
% =======================================================================================================

\author{Christoph Aymanns\thanks{QuantCo. Email: \href{mailto:christoph.aymanns@gmail.com}{christoph.aymanns@gmail.com}.}
\and
Jakob Foerster\thanks{University of Oxford. Email: \href{mailto:jakob.foerster@eng.ox.ac.uk}{jakob.foerster@eng.ox.ac.uk}.}
\and
Co-Pierre Georg\thanks{Frankfurt School of Finance and Management. Email: \href{mailto:c.georg@fs.de}{c.georg@fs.de}.}
\and
Matthias Weber\thanks{University of St.\ Gallen and Swiss Finance Institute. Email: \href{mailto:matthias.weber@unisg.ch}{matthias.weber@unisg.ch}.}}

% \date{\today}
\date{}
\maketitle
\thispagestyle{empty}

% \vspace{-1cm}

% =======================================================================================================
%
\begin{abstract}\noindent 
We propose multi-agent reinforcement learning as a new method for modeling fake news in social networks. 
This method allows us to model human behavior in social networks both in unaccustomed populations and in populations that have adapted to the presence of fake news. In particular the latter is challenging for existing methods. 
We find that a fake-news attack is more effective if it targets highly connected people and people with weaker private information.
Attacks are more effective when the disinformation is spread across several agents than when the disinformation is concentrated with more intensity on fewer agents. 
Furthermore, fake news spread less well in balanced networks than in clustered networks.
We test a part of our findings in a human-subject experiment. The experimental evidence provides support for the predictions from the model, suggesting that the model is suitable to analyze the spread of fake news in social networks.
% [145 words]

%Over the last few years deep multi-agent reinforcement learning (DMARL) has become an increasingly active area of research with hundreds of papers submitted to top machine learning conferences every year. However, so far there have been few real world use cases that benefited from the progress in the field. 
%We use DMARL to develop a flexible computational model of fake news on social networks in which agents act according to learned best response functions. We achieve this by extending an information aggregation game to allow for fake news and by representing agents as recurrent deep Q-networks (DQN). In the game, agents repeatedly guess whether a claim is true or false taking into account an informative private signal and observations of actions of their neighbors on the social network in the previous period. We incorporate fake news into the model by adding an adversarial agent, the attacker, that either provides biased private signals to, or takes over, a subset of agents. The attacker can follow either a hand-tuned or trained policy. Our model allows us to tackle questions that are analytically intractable in fully rational models, while ensuring that agents follow reasonable best response functions. Our results highlight the importance of awareness, privacy and social connectivity in curbing the adverse effects of fake news and open an entire new real world application area for DMARL.
\end{abstract} \vfill
%
% =======================================================================================================

% =======================================================================================================
%
\section{Introduction}\label{sec::introduction} % (fold)
%
% =======================================================================================================
\setcounter{page}{0}
%\paragraph{Problem setting}
Fake news is spread on social networks to manipulate users' perceptions of facts. The 2016 and 2024 US presidential elections highlighted the threat that fake news poses to open societies. Similarly, Brexit, COVID-19, and the war in Ukraine provide recent contexts in which fake news has played a crucial role. However, relatively little is known about the spread of fake news on social networks, primarily because decision-making in these networks is highly complex and not well captured by existing models.

In this paper, we use deep multi-agent reinforcement learning to approximate the solution of an information aggregation game on social networks. This approach enables richer learning dynamics, though it is unclear whether it represents a more realistic model of human behavior. To address this, we conduct a controlled experiment, showing that the predictions of our model regarding the spread of fake news align with observed human behavior in a laboratory setting.

We model the spread of fake news as a sequential information aggregation game involving $N$ agents interacting through a predetermined social network. Agents are presented with a claim that is either factually true or false (the "state of the world") and decide on a binary action. They are rewarded if their action aligns with the claim's veracity. Each agent receives a private signal about the state of the world and, at each time step, observes the previous actions of their neighbors.

The game is subtle because choosing the action that best matches an agent's posterior information is not always optimal; such choices may lead other agents to imitate them, revealing no new information. This subtlety makes the game ideal for independent Q-learning (IQL), which we use to computationally solve the game. Our approach is flexible, computationally tractable, and provides several key insights.

First, we establish a baseline condition of information aggregation without an attack. We show that IQL achieves higher accuracy in information aggregation than the \mbox{DeGroot} learning benchmark. Second, our numerical analysis examines factors that influence the effectiveness of the attack, such as the position of the target's network, the global structure of the network, and the distribution of biases among agents. Attacks targeting highly connected agents or those with weak private signals are more effective. Clustered networks are more susceptible to attack than balanced networks, and distributed attacks are more effective than concentrated ones. Third, when attackers are present during the training phase, citizens learn to adapt, resulting in improved accuracy in information aggregation. Lastly, we explore scenarios where an attacker knows the true state of the world and acts to mislead citizens. While simple attack strategies are quickly countered by citizens, attacks executed by a separate deep Q-learner are significantly more effective.

Although simulations provide valuable predictions, they are insufficient to determine whether independent Q-learning realistically models human behavior in information aggregation games. To address this, we test our model's predictions through a laboratory experiment involving six treatments and 560 participants. Participants, organized into independent networks of eight, are tasked with correctly guessing the state of the world over several periods. As in our model, participants receive private information and observe the actions of their connected neighbors. We employ a $2\times 3$ factorial between-subjects design, varying two types of networks and three types of attacks. Overall, the experimental results align with the model's predictions.

%
% LITERATURE REVIEW
%
Our work relates to several strands of literature. First, it builds on foundational models of information aggregation in social networks \citep{BikhchandaniHirshleiferWelch1992,Banerjee1992,bala1998learning}, where agents attempt to infer an unknown state of the world using private signals and observed actions of their neighbors.\footnote{This literature is itself part of the literature on social and economic networks more generally, not necessarily focusing on information diffusion \citep[e.g.,][]{jackson2002evolution,jackson2007diffusion,galeotti2010network}.} Recent surveys highlight both rational and heuristic approaches \citep{bikhchandani2024information,Golub2016}. Typically, however, the standard setting abstracts away from misinformation or malicious attacks. Rational models focus on conditions under which agents' beliefs converge to the truth in large populations and over long time horizons \citep{mossel2015strategic}, but these baseline models are often difficult to extend to finite horizons or more complex environments.% this part above added for top-5 submssions: ,bala1998learning AND THE FOOTNOTE WITH THE GENERAL ECON LITERATURE (INCLUDING THE YARIV PAPERS)

Recent contributions have begun to explore richer settings. For example, some models consider learning about a changing state and how individuals adapt to evolving conditions \citep{dasaratha2023learning}, some works examine the interplay between informational herding, experimentation, and contrarianism \citep{smith2021informational}, while others analyze how even fully rational agents can exhibit groupthink \citep{harel2021rational}. Work on learning in networks also incorporates heterogeneous and misspecified models, addressing questions of robustness and the conditions under which learning may fail or be distorted \citep{bohren2021learning}. Related research provides insight into the dynamics of learning in networked environments and how the structure of interactions affects equilibrium outcomes \citep{board2021learning}.

Heuristic or boundedly rational models \citep[e.g.,][]{DeGroot1974,GolubJackson2010} offer flexibility in capturing transient behavior and changing network conditions, but they may oversimplify agent responses or fail to account for strategic manipulation. By contrast, our approach incorporates complexity and adaptability---allowing for dynamic, strategic interactions, and systematic misinformation---while remaining computationally tractable.% this part above added for top-5 submssions: ,GolubJackson2010 ; removed in same place Golub2016, (which probably should have been the GolubJackson paper in the first place that was already in the reference.bib file)

A growing body of work now examines the role of misinformation and fake news in shaping information aggregation. Empirical and theoretical studies document how misinformation can spread rapidly, influencing political outcomes and social beliefs \citep{allcott2017social,vosoughi2018spread,acemoglu2019fake}. Some have begun to model and quantify the strategic introduction of fake news \citep{papanastasiou2019fake}, emphasizing that traditional learning models without adversaries may underestimate the complexity of information flows in networks. In contrast, our approach employs deep multi-agent reinforcement learning to flexibly approximate near-optimal policies in scenarios where agents face both uncertainty and potential manipulation. In doing so, we bridge the gap between rational but analytically complex models and simple heuristic frameworks that do not adapt to strategic deception. Our method thus allows the study of transient behavior under fake news attacks, complementing existing analyses in the literature on strategic information diffusion \citep{galeotti2009influencing,alon2010note,galeotti2013strategic,etesami2014complexity}.
% this part above added for top-5 submssions: galeotti2009influencing, galeotti2013strategic,

Second, our work relates to the literature on deep multi-agent reinforcement learning (DMARL) as a method for analyzing complex strategic problems with multiple interacting agents. DMARL leverages deep neural networks to approximate policies or value functions, enabling the study of environments that are otherwise intractable under classical solution concepts. A number of surveys highlight the rapid evolution of this field and its diverse applications, including challenges related to non-stationarity, partial observability, and coordination \citep{DuDing2021, NingXie2024, OroojlooyHajinezhad2023}. Early contributions focused on developing algorithms that could foster cooperation or stable outcomes in relatively contained settings. More recent work demonstrates the potential of DMARL to achieve expert-level performance in complex strategic domains \citep{vinyals2019grandmaster, baker2019emergent, foerster2018learning, crandall2018cooperating} and to model how agents adapt in dynamic, uncertain environments. Within the context of economics and management, DMARL techniques have informed studies on misinformation diffusion, strategic pricing, and capacity management under uncertainty \citep{papanastasiou2019fake,MolaviTahbaz-SalehiJadbabaie2017}, extending and enriching traditional theoretical frameworks \citep[e.g.,][]{mossel2015strategic}. In this paper, we employ Independent Q-Learning \citep{tampuu2017multiagent}, a straightforward DMARL approach that balances computational tractability with the ability to capture nuanced, adaptive strategies in the presence of fake news.

Third, our paper is related to the experimental literature studying human interaction and learning in networked environments. This body of research includes experiments on cooperation and public good provision in networks \citep{vanLeeuwen2019centrality,vanLeeuwen2020competition}, on bargaining in networked markets \citep{agranov2021commitment}, as well as on how information and behavior spread through social connections \citep{centola2010spread,goyal2017information}. A growing subset of this literature directly investigates how individuals learn from and respond to information in networked settings, including how they react to false or biased signals. For instance, \cite{chandrasekhar2020testing} test models of social learning on networks using a field experiment, \cite{corazzini2012false} examine the impact of false information feedback on learning and decision-making, and \cite{muellerfrank2013nonbayesian} analyze non-Bayesian updating in experimental social learning contexts. Other studies focus on dynamic or changing environments, showing how individuals continuously update their beliefs and actions \citep{grimm2018experiment}. Notably, there are still relatively few human-subject experiments that address fake news or deliberate disinformation directly. A notable exception is \cite{stewart2019information}, who investigate a network voting game with bots. In contrast to that study, we focus on information aggregation more generally and tie our experimental findings to the predictions of a flexible reinforcement-learning model.
% this part above added for top-5 submssions: , on bargaining in networked markets \citep{agranov2021commitment}

%
% CONTRIBUTION
%
We develop a flexible computational model of fake news on social networks in which agents act according to learned best response functions. Furthermore, we provide evidence that the predictions of this model are realistic in the sense that they correspond to findings from a human-subject experiment. Our analysis helps us to answer three main questions: (i) What determines the effectiveness of the attacker in reducing the accuracy of information aggregation? (ii) How well can agents learn to adapt to the presence of an attacker? And (iii) When the attacker and agents learn simultaneously, how does behavior evolve over time? 

Concerning the first question, we find that fake news spreads more easily in networks with separate clusters than in more "balanced" networks. Furthermore, a fake news attack is more effective if the attacker spreads less careful disinformation over more agents than if he invests his efforts in more thoroughly convincing fewer people. Fake news attacks are also more effective if agents with more connections on the network are attacked and if agents are attacked who have weaker private information. Concerning the second question, we find that agents can indeed learn to adapt to the presence of fake news. However, learning to adapt to fake news comes at a cost---if agents are trained in the presence of fake news, the accuracy when no fake news are present is lower than in the case that no fake news are possible at any stage. Regarding the third question, when agents and attackers learn simultaneously, behavior does not necessarily converge, but interesting irregular cyclical dynamics may evolve. Our findings have important implications for the fight against disinformation, which we discuss in the concluding section.

% =======================================================================================================
%
% <Chris/Jakob Start>
\section{Model of Fake News} %This is \textit{formal / abstract} description of the setting
\subsection{Background: Information Aggregation as a Network Game} %bookwork -- not fake news specific
\label{sec:background}
% Modeling Fake News as a Network game 
We model information aggregation as a network game over $T$ periods indexed by $t=1, \dots, T$ with a set of agents $N$ (for details, see \citealp{mossel2015strategic,Golub2016}). Agents interact via a fixed, directed, graph $\net(N, E)$, where $E$ denotes the set of edges connecting the nodes $N$. For concreteness, one can think of the agents as users of a social network, such as Twitter/X. A directed link from agent A to agent B would then imply that A follows B on Twitter/X.

Agents are presented with a claim that is either \textit{true} or \textit{false}. Again, for concreteness, one can think of the claim as a factual statement encountered in a piece of news that has been shared among the agents on the social network. 
% <<
We restrict our analysis to the case when claims are factual statements that can be objectively verified as true or false. While this prevents us from studying the dynamics of opinions, our focus in on the spread of fake news.
% >>
True claims are denoted as $\theta=1$ and false claims as $\theta=0$. Agents receive private signals $s^i \sim F_\theta$ once at the beginning of the game. In the baseline condition (i.e., no attack) $F_\theta$ is such that private signals are always informative but noisy. In each period each agent chooses a binary action $\act^i_t \in \{0, 1\}$ and observes the actions of their neighbors on the graph from the previous period. At the end of the game (i.e., after $T$ periods), agents are rewarded for actions that matched the veracity of the claim. For a given claim $\theta$ and sequence of actions $\{\act_t^i\}$, an agent's total discounted payoff is $R_i = \sum_t^T \gamma^t \id\{a_t^i = \theta\}$, where $\id\{a_t^i = \theta\}$ is agent $i$'s payoff for period $t$.

%Note that in this game, for $t<T$, it is not necessarily optimal to choose the action that matches the maximum of the posterior likelihood of $\theta$. This is because an agent may want to prevent others from simply copying its action, since that reveals no information about the others' private signal. Injecting noise into actions may therefore be beneficial. Despite these concerns, it can be shown that, under certain conditions on the network, in the limit $T \to \infty$ (number of periods) and $n \to \infty$ (number of agents), optimal agents will converge and will agree on the correct $\theta$ (\citealp{mossel2015strategic}).

\subsection{Modeling the Spread of Fake News} % our extension of the formal model
We extend the model of information aggregation by introducing an adversary---the so called \textit{attacker}---who wants to reduce the other agents' total payoff $\sum_i R_i$. To disambiguate between the attacker and other agents, we refer to other agents as \textit{citizens}. The attacker wants to persuade citizens to support false claims. 
We model a \textit{biased signal attack}, where the attacker distributes a fixed budget of bias $\beta = \sum_i \beta^i$ across a subset of the agents such that citizen $i$'s private signal is drawn from $s^i \sim f_\theta(\beta_i)$.
Specifically, $f_\theta(\beta_i) = \Norm(\theta + \beta^i (1 - 2 \theta), \sigma^2)$; that is, we move the mean of the distribution for the signal of citizen $i$ \textit{away} from the truth, $\theta$, by a factor of $\beta^i$.  %Second, in the \textit{agent takeover attack}, the attacker can directly take over an agent and act on his behalf. In this case the attacker optimizes his attack policy by independent Q-learning.
% << was a footnote
Biased signals can be thought of as targeted ads presented to the users of a social network. Indeed, it has been shown that fake news attacks have taken this form \citep{chiou2018fake}, where the bias budget corresponds to an attacker with a limited budget for targeted social media ads.
% >>

% At first, one possible approach might be to try to solve this model using standard game theoretic approaches. However, the complexity of the fake news setting is beyond the scale of these methods at the current time. 

\section{Simulation Approach to Studying the Model} %our method
% =======================================================================================================
We model agent behavior using deep multi-agent reinforcement learning (MARL), with actions derived from policies optimized through this approach. This method balances full rationality and bounded rationality with heuristic-based decisions, enabling the study of agent adaptation to fake news and facilitating qualitative predictions about specific fake news attack scenarios. MARL has been shown to scale to complex multi-agent settings and partial observability \citep{foerster2018learning}.  

Specifically, our approach is as follows: We separate our study into a \textit{training} and \textit{testing} phase. 
At the beginning of the training phase we initialise agents with a policy represented as a deep neural network, which maps from the information state of each agent to a distribution over actions. During \textit{training} we simulate many games (``episodes'') and optimise the policies of all agents after each episode to improve individual payoffs (``rewards''). Training continues until a maximum number of episodes is reached, which we tune to achieve policy convergence. During the \textit{testing phase} we instead keep the final policy parameters from the training phase fixed and repeatedly simulate the game to evaluate the performance under a range of different \textit{attack scenarios}. In the following we provide further details of our method: we first describe single-agent, then multi-agent RL and finally connect these general methods to our model.

\paragraph{Single-agent RL:} For ease of introduction, we first consider \textit{fully observable, single agent RL}; that is, a Markov Decision Process (MDP) specified by the following tuple: ${\langle}S, A, P, r,
\gamma{\rangle}$. The environment is described by a Markov state, $s \in
S$, and at each time step the agent chooses an action $\act \in A$, which induces a (probabilistic) transition to the next state via the state transition function $P(s'|s,a): S \times A \times S \rightarrow [0,1]$. The reward function assigns an \textit{instantaneous reward} to each state-action, successor state tuple, $r(s,a, s'): S \times A  \times S
\rightarrow \mathbb{R}$. $\gamma \in [0,1)$ is a discount factor.

The task of the agent is to maximize the discounted reward per episode, $R_t = \sum_{l=0}^\infty \gamma^l r_{t+l}$. The agent's \textit{policy} is a mapping from states to actions, $\pi(a|s): S \times A  \rightarrow [0,1]$,  and induces a state-action-value function  (Q-function), $Q^{\pi}(s_t, a_t) = \exs{s_{t+1:\infty},a_{t+1:\infty}}{R_t|s_t,a_t}$, i.e.\ the expected forward looking return after taking action $a$ in state $s$. Q-learning \citep{SuttonBarto1998} aims to estimate the optimal action-value function,
$Q^{*}(x,a) = \max_{\pi} Q^{\pi}(x,a)$, via an estimated Q-function, $Q(s,a)$.

Training is carried out by collecting \textit{samples} from the environment in order to obtain Monte-Carlo estimates of this expected return. For any sampled transition, the current estimate, $Q(s,a)$,  is compared to a greedy one-step lookahead using the Bellman optimality operator, $\mathcal{T}Q(s,\act) = \E_{s'} \left[ r +  \gamma  \max_{a' } Q(s',a') \right]$. We use samples to evaluate the Bellman update:
\[
Q(s,a)_{k+1} =  Q(s,a)_{k} + \alpha \left( r +  \gamma  \max_{a' } Q(s',a')_{k}  - Q(s,a)_{k} \right).
\]
Here $k$ is the iteration number, $\alpha$ is the learning rate and $ r +  \gamma  \max_{a' } Q(s',a')_{k}  - Q(s,a)_{k} $ is commonly referred to as the temporal-difference or TD-error. In the tabular case the Q-values for each state-action pair are maintained separately and the Bellman update is a contraction mapping. As a consequence, under mild conditions,  at convergence this iterative process results in the optimal $Q$-function, $Q(s,a)_{\infty} =Q^{*}(s,a) $. Finally, $Q^{*}(s,a)$  trivially defines the optimal policy $\pi^*(x,a) = \delta(\arg\max_{a'} Q^*(x,a')- a)$, where $\delta(\cdot)$ is the Dirac-delta function. In contrast to the tabular case, recurrent deep Q-networks (DQN) use a neural network parametrized by a large number of weights, $\phi$, to represent the $Q$-function \citep{Mnih2015}.

To reduce the variance of the update process, the average square of the TD-error across a large number of transitions (the \emph{batch}) is used as the DQN-loss. Using backpropagation, the parameters of the neural network are updated to minimize the magnitude of the DQN error $\mathcal{L}(\phi) = \sum_{j=1}^b [( y_j^{DQN} - Q(x_j,\act_j; \phi))^{2}]$. Here $y_j^{DQN} =  u_j + \gamma \max_{a_j'} Q(x_j',a_j';\phi^{-})$ is the \textit{target function} and $\phi^{-}$ is the target network, which contains a stale copy of the parameters. This target network helps to stabilize the training.

% Partially observable RL
So far we have assumed that the agent has access to the Markov state, $s$, of the system. In a partially observable setting we augment the MDP with an \textit{observation space}, $Z$, and observation function $O(s)$. In particular, the observation $z \in Z$ is produced by the observation function $O(s): S  \rightarrow Z$. We further define an action-observation history $\tau \in T \equiv (Z \times A)^{*}$, which is  used to condition a stochastic policy $\pi(a|\tau): T \times A \rightarrow [0,1]$. In recurrent deep RL \citep{HausknechtStone2015}, this is commonly achieved using recurrent neural networks, such as LSTMs \citep{HochreiterSchmidhuber1997} or GRUs \citep{cho2014properties}, which we also use here. % \citep{chung2014empirical}

% Partially observable MARL. 
\paragraph{Multi-agent RL:} Since our model of fake news involves a number of different agents with different observations, we use partially observable MARL.
Here, each agent $i \in N$ receives a private, partial or noisy observation $O(s, i)$ of the Markov state, where $i$ is the agent index and $O$ is the observation function. Each agent also receives individual reward $r^i_t$ and takes actions $a^i_t$. Furthermore the state-transition conditions on the joined action $\mathbf{a} \in \mathbf{A} \equiv A^{n}$. Specifically, we use independent Q-learning, where each agent estimates an individual $Q$-function $Q^i(\tau^i,\act^i)$, treating other agents and their policies as part of a \textit{non-stationary environment}. Following standard practice, we further use parameter-sharing across agents combined with an agent specific index, $i$, in the observation function to accelerate learning while still allowing for specialization of policies.

%Training 
\paragraph{Our model:} We can now specify our model as a MARL task. For a \textit{given} episode, the environment consists of (i) the social network, (ii) a realization of the state of the world $\theta$, (iii) a realization of the bias vector, $\mathbf{\beta}$, and (iv) a realisation of all private signals. Here all random variables are realised according to the distributions in Section~\ref{sec:background}.

At time step $t$ \textit{within} an episode, each citizen observes their private signal $s^i$, her last action and the last actions of her neighbors $\{a^j_{t-1}\}_{j \in B_i}$ ($B_i$ is the set of neighbors of agent $i$ augmented by $i$) and the agent id $i$, which identifies the agent's network position. The set of all last actions $a^i_t$ and private signals together with the true state of the world, $\theta$, and the attack vector $\beta$, form the Markov-state of the environment. The reward function for agent $i$ is simply their utility function specified above, $r^i_t = \id(a^i_t, \theta)$. We illustrate this process in Algorithm \ref{algo:simulation}.

During the model training phase, many game episodes are simulated as described above, and after each episode the policy parameters are updated using independent Q-learning. This process is described in Algorithm \ref{algo:training}. During the model testing phase, we simply evaluate a trained policy using Algorithm \ref{algo:simulation}.

\begin{algorithm}[H]
\caption{Game simulation}
\label{algo:simulation}
\KwIn{
\begin{itemize}
        \item Social network $\net(N, E)$.
        \item Policy parameterized by $\phi$: $\pi_\phi$.
        \item Draw state of the world $\theta$\;
        \item Draw bias vector $\mathbf{\beta}$ (if applicable)\;
        \item Draw private signals $s_i \sim f\theta(\beta^i)$\;
        \item Initialize $a^i_{-1}=1$\;
    \end{itemize}
}
\KwOut{
    History of observations, actions and rewards.
}
\For{period $t$ within the episode}{
        \For{each agent $i$}{
            Construct observation $o^t_i = (s^i, a^i_{t-1}, \{a^j_{t-1}\}_{j \in B_i}, i)$\;
            Draw action from policy $a^i_t \sim \pi_\phi(o^t_i)$\;
            Calculate reward: $r^i_t = \id(a^i_t, \theta)$\;
        }
}

\end{algorithm}

\begin{algorithm}[H]
\caption{Simplified model training}
\label{algo:training}
\KwIn{
\begin{itemize}
        \item Class of social networks $\mathcal{C}$, see Section \ref{simulation_setup}
        \item Training scenario (with or without bias attack), see Section \ref{simulation_setup}.
        \item Initial policy parameterized by $\phi$: $\pi_\phi$.
        \item Number training episodes $K$.
    \end{itemize}
}
\KwOut{
    Trained policy $\pi_\phi$.
}
\For{each training episode}{
    Draw a social network from $\mathcal{C}$;
    Draw bias vector $\mathbf{\beta}$\;
    Draw private signals $s_i \sim f\theta(\beta^i)$\;
    Initialize $a^i_{-1}=1$\;
    \For{period $t$ within the episode}{
        Run game simulation\;
        Collect history of observations, actions and rewards\;
        Update $\phi$ using independent Q-learning\;
    }
}
\end{algorithm}

%To maintain a constant length input vector, unobserved actions are encoded as $-1$. Thus, the corresponding DQN is a function $Q : \R \times \{-1, 0, 1\}^{|N|} \times N \times \{0,1\} \mapsto \R$.

%DANGER END

%Testing
% In the \textit{biased signal attack}, the attacker is \textit{not} represented by an RL policy. Instead, the biases are hand-tuned. At the beginning of each episode the attacker chooses a set of citizens uniformly at random from $N$ and delivers biased signals to them as outlined above.

% find a spot for this
% Our objective is to improve our understanding of the propagation of fake news in social networks and discover potential interventions to curb its effectiveness. This raises three main questions: (i) What determines the effectiveness of an attacker in reducing the accuracy of information aggregation? In particular, how does the network structure and the choice of the attacked agent affect attack effectiveness? (ii) How well can agents learn to adapt to the presence of an attacker? (iii) When attacker and citizens learn simultaneously (agent takeover attack), how do policies change over time? To study these questions we set up a series of training and testing scenarios.

\subsection{Measuring and Comparing Information Aggregation }
Intuitively, agents should be able to learn to extract information from their private signals and their neighbors' actions, such that at the end of an episode their expected reward ($\E[r^i_T]$) should be significantly higher than at the beginning of an episode ($\E[r^i_0]$). That is, agents should aggregate information over time. In the following, we give a formal definition of information aggregation for finite $N$ and $T$ that is useful in our case.
\begin{definition}[Information aggregation]
Let $\mathcal{F} = \{s^i\}_{i\in N}$ denote the set of all realized signals in a given episode. Let $\hat{\theta} = \text{argmax}_\theta \Prob[\theta \mid \mathcal{F}]$ denote the maximum-a-posteriori estimator of $\theta$ given $\mathcal{F}$. Then information is aggregated (perfectly) if for all $i \in N$, we have $a_T^i = \hat{\theta}$.
\end{definition}
That is, under this definition information is aggregated if agents act as if they had seen all private signals. In reality, it is not reasonable to expect perfect information aggregation with actions learned via independent Q-learning. We therefore define the following measure of information aggregation.
\begin{definition}[Accuracy of information aggregation]
The accuracy of information aggregation at time step $t$ is $A_t = \E[\id\{a_t^i = \theta\}]$.
\end{definition}
A useful baseline to compare $A_t$ to is of course $\E[\id\{\hat{\theta} = \theta\}]$. Another useful baseline is $\E[\id\{\tilde{\theta} = \theta\}]$, where $\tilde{\theta} = \text{argmax}_\theta \Prob[\theta \mid s^i]$ is the MAP estimator given only a single private signal. $A_T$ can never exceed $\E[\id\{\hat{\theta} = \theta\}]$ and $A_1$ can never exceed $\E[\id\{\tilde{\theta} = \theta\}]$ and of course $\E[\id\{\hat{\theta} = \theta\}] > \E[\id\{\tilde{\theta} = \theta\}]$ provided $|N| > 1$. Once we have trained the agents via Q-learning, it is easy to estimate $A_t$ by evaluating a set of batches of games $B$ and computing $\hat{A}_t = 1/|B| 1/|N| \sum_{b \in B, i \in N} \id\{a_{t, b}^i = \theta\}$,
% \begin{equation}\label{EQ::accuracy}
% 	\hat{A}_t = \frac{1}{|B|} \frac{1}{|N|} \sum_{b \in B, i \in N} \id\{a_{t, b}^i = \theta\},
% \end{equation}
where $a_{t, b}^i$ is the action taken by agent $i$ in batch $b \in B$. In the following we will simply refer to $\hat{A}_t$ as the accuracy.

Lastly, we compare the accuracy of the DQN against the DeGroot heuristic \citep{DeGroot1974}. Under this heuristic information aggregation rule at $t=0$, we set $a^i_0 = \tilde{\theta}(s^i)$. In each subsequent period actions are set to $a_t = U a_{t-1}$, where $a_t$ is the vector of actions taken by citizens and $U = D^{-1} (I + M)$, where $I$ is the identity matrix, $M$ is the graph adjacency matrix and $D$ is a diagonal matrix with entries $D_{ii}$ being the number of neighbors of node $i$ plus one (i.e., a self loop). Information is therefore aggregated through repeated averaging of past actions.

\subsection{Simulation Setup and Parameters}
\label{simulation_setup}
 For simplicity we model the unconditional probability of a claim being true or false as $\Prob(\theta=0)=\Prob(\theta=1)=1/2$. We take $\sigma^2=1$ for the variance of the private signals.

 % -Network Distribution
For the simulations, we consider three types of \textit{undirected} graphs, a Barabasi-Albert random graph \citep{barabasi1999emergence}, a clustered graph of fully connected nodes, and a balanced graph obtained by randomly rewiring the clustered graph.
% , see Figure \ref{FIG::networks} for an illustration. 
For the illustrations, we use a Barabasi-Albert graph with $|N| = 10$, a clustered graph with three clusters ($n_c=3$) of fully connected nodes with four nodes each ($s_c=4$) such that $|N|=12$, and a balanced graph obtained by randomly rewiring the clustered graph ($|N|=12$). While we use these specific graphs for the illustrations in the paper, note that the results that we show also hold for other network sizes.

Random rewiring involves repeatedly selecting two edges uniformly at random and swapping the terminal nodes of these edges. This preserves the degree sequence, that is the number of neighbors of each node, but it removes the clusters, so that the rewired graph is ``balanced''. The Barabasi-Albert random graph is often used to model social networks and is therefore a natural choice for our application. The clustered graph is a stereotypical example of a social network which is comprised of cliques that are weakly connected. It thereby represents a more ``fragmented'' or ``polarized'' social network. The balanced graph removes these clusters from the graph while maintaining the degree sequence and can thus help isolate the effect of clusters on information aggregation. We analyze different types of networks, but the network remains fixed during each game.

\begin{figure}[htbp]
\begin{center}
%\framebox[4.0in]{$\;$}
\includegraphics[width=1.03\textwidth, trim= 2cm 0 0 0, clip=false]{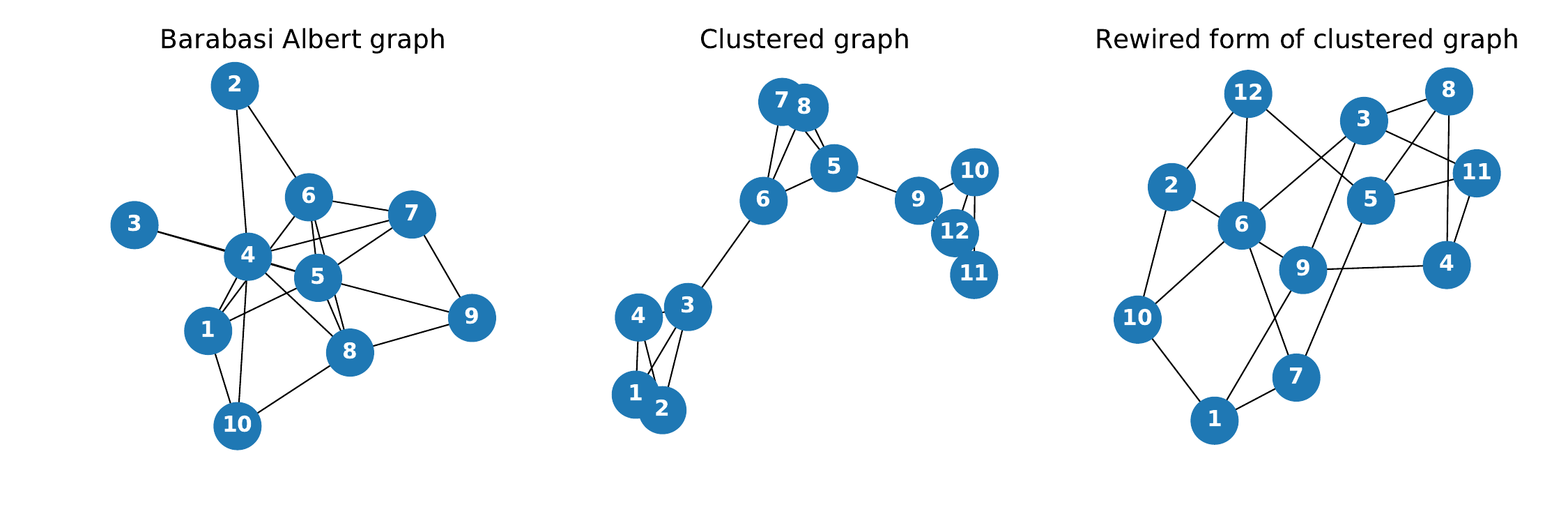}
\end{center}
\caption{Left: Instance of a Barabsi-Albert random network with $|N| = 10$ that was used in our analyses (unless otherwise stated). Middle: stereotypical clustered network with $|N|=12$ with three ($n_c=3$) fully connected clusters of four nodes ($s_c=4$) each. Right: the clustered network in balanced form produced by random rewiring of link pairs (preserves degree distribution). These networks were used in our analysis of the effectiveness of information aggregation and attack (i) in clustered vs.\ balanced networks and (ii) for spread out vs.\ focused attack strategies.}\label{FIG::networks}
\end{figure}

% <<
Figure \ref{FIG::networks} illustrates the networks used in this paper. 
% >>

% -Train / test setup
\label{par:training_scenarios}
We implement the following training scenarios.\footnote{Our neural network architecture implements the IQL architecture in \cite{foerster2016learning}.} (i) \textbf{Baseline:} We train citizens in the absence of any attack; that is, all citizens receive unbiased signals and no citizen is taken over by the attacker. %We train baseline models for the Barabasi-Albert, clustered and balanced graphs. 
This scenario establishes a benchmark for the ability of IQL agents to learn to aggregate information in the absence of attack. (ii) \textbf{Biased signal attack:} We train citizens in the presence of a biased signal attack where one or more randomly chosen citizens receive a biased signal. % single one: ($\beta=3$).
%We train this model for the Barabasi-Albert graph only. 
This scenario allows us to evaluate whether citizens can learn to adapt to the presence of a biased signal attack and thereby mitigate the adverse effects of the attack.

% \subsection{Testing scenarios} % (fold)
\label{par:testing_scenarios}
Given the models obtained for these different training scenarios we conduct a number of testing scenarios. First, for each attack training scenario, we evaluate accuracy, as defined above, both in the presence of the corresponding attack (as trained) and in the absence of any attack. For the baseline scenario we consider two testing scenarios in addition to testing in the absence of any attack (as trained). First, we evaluate accuracy in the baseline model for the Barabasi-Albert graph in the presence of a biased signal attack with a single attacked agent and $\beta=3$. Second, we evaluate accuracy in the baseline model for the clustered and balanced graphs under two biased signal attack scenarios. In the \textit{focused} scenario, a single, randomly chosen citizen receives a biased signal with $\beta^i=3$. In the \textit{spread} scenario, two, randomly chosen citizens receive biased signals with $\beta^i=1.5$ each.

% <<
% =======================================================================================================
%
\section{Simulating the Spread of Fake News Using Independent Q-learning}\label{sec:results}
%
% =======================================================================================================
% >>

Our analyses yield four main results, which we will discuss in more detail below. 

\subsection{Baseline Information Aggregation} 
\begin{figure}[htbp]
\begin{center}
%\framebox[4.0in]{$\;$}
\includegraphics[width=0.9\textwidth, trim= 0 0 18cm 0, clip=true]{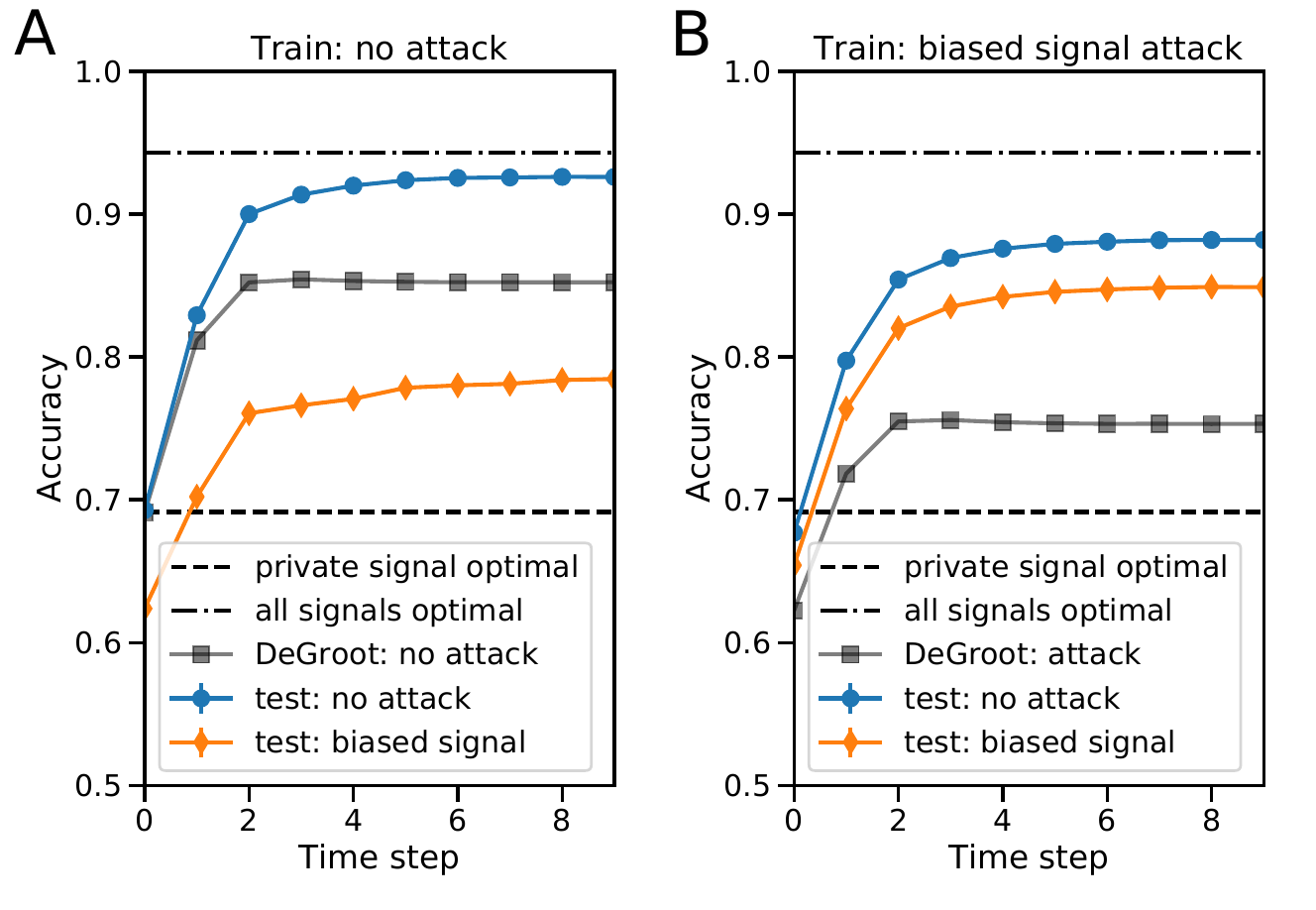}
\end{center}
\caption{(A-B): Information aggregation over time in a Barabasi-Albert graph. (A) Baseline scenario. Under the attack test scenario, a single agent receives a biased signal. (B) Biased signal attack scenario. The upper and lower dashed lines corresponds to the benchmarks $\E[\id\{\hat{\theta} = \theta\}]$ and $\E[\id\{\tilde{\theta} = \theta\}]$ respectively. Square markers correspond to the accuracy achieved by the De Groot information aggration heuristic in the baseline and biased signal attack scenarios respectively.}\label{FIG::combined_plot}
\end{figure}

In the absence of an attack, citizens learn to aggregate information with an accuracy close to the optimal benchmarks $\E[\id\{\hat{\theta} = \theta\}]$ as $t\to T$ and $\E[\id\{\tilde{\theta} = \theta\}]$ for $t=0$. Our method's accuracy substantially outperforms the accuracy achieved by the DeGroot information aggregation heuristic. 
%This highlights the appropriateness of IQL to solve the standard information aggregation game. 
In the presence of an attack, information aggregation is severely disrupted. These results are illustrated in Figure \ref{FIG::combined_plot} (A).

\subsection{Determinants of the Effectiveness of an Attack} 
Our simulations allow us to study the following determinants of attack effectiveness: the network position of the attacked agent, the signal strength of the attacked agent, the network structure (clustered vs.\ balanced) and the distribution of bias across agents (focus vs.\ spread). We consider each in turn.

\begin{figure}[htbp]
\begin{center}
%\framebox[4.0in]{$\;$}
\includegraphics[width=0.9\textwidth,  trim= 22.5cm 0 0 0, clip=true]{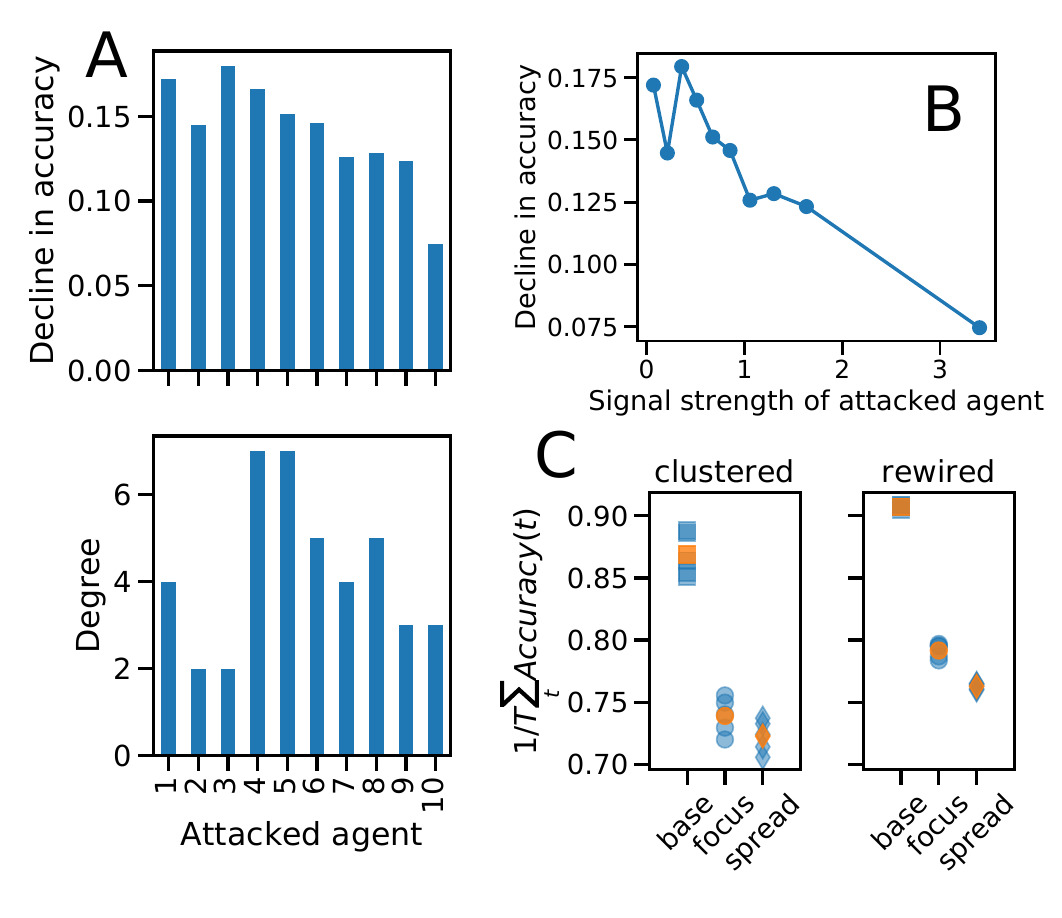}
\end{center}
\caption{(A-B) Agents are trained in the absence of any attacker on a Barabasi-Albert graph. (A) Top: Average decline in accuracy conditional on agent id which determines network position. Bottom: degree of attacked agent. (B) Effectiveness of conditioning on attacked agent signal strength. (C) Agents trained in the absence of any attacker for clustered (left) and balanced (right) graphs. Base: no attack. Focus: a single, randomly chosen agent receives a strong biased signal ($\beta^i=3$). Spread: two, randomly chosen agents receive weak biased signals ($\beta^i=1.5$). Blue markers represent training runs with different seeds for a given network and attack scenario. Orange markers are averages over the different training runs. Error bars computed over outcomes with different random seeds but fixed neural network weights are too small to be visible on the graphs.}\label{FIG::combined_plot_2}
\end{figure}

To study the effect of the network position, we compute the accuracy in the final time step ($t=T$) conditional on the network position of the attacked agent when evaluating the baseline model under the biased signal attack in the Barabasi-Albert network. Denote this conditional accuracy by $\hat{A}_T(i)$, where $i$ is the attacked agent. Let $\hat{A}_T$ denote the baseline accuracy in the absence of an attack. We then define the decline in accuracy relative to the baseline as $\Delta A(i) = \hat{A}_T - \hat{A}_T(i)$. We plot $\Delta A(i)$ as a function of $i$ in Figure \ref{FIG::combined_plot_2} (A). It is clear that there is substantial heterogeneity in the attack effectiveness. As the tight correlation between $\Delta A(i)$ and the degree of $i$ in Figure \ref{FIG::combined_plot_2} (A) shows, the heterogeneity in $\Delta A(i)$ arises from the network structure, which implicitly gives more influence to those nodes with more neighbors (higher degree). If an attacker is able to target the biased signal to such highly connected nodes, the attack will be more effective.

Next, we study the effect of the signal strength. Again, we restrict ourselves to evaluating the baseline model under the biased signal attack in the Barabasi-Albert network. A natural measure of a signal's strength, or informativeness, is the absolute value of the log likelihood ratio of the two states of $\theta$ conditional on the signal. Let $f(\theta \mid s)$ denote the posterior distribution over $\theta$ given the signal $s$. 
Then, the signal strength is given by $L(s) = \vert \log [f(\theta = 1 \mid s) /f(\theta=0\mid s)] \vert$. 
Let $L_k$ denote the $k$th decile of the empirical distribution of signal strengths. We define the conditional accuracy $\hat{A}(L_k)$ as the average accuracy in the final time step conditional on the attacked agent's signal strength lying in $[L_k, L_{k+1}]$. $\Delta A(L_k)$ is defined analogously to $\Delta A(i)$. We plot $\Delta A(L_k)$ against $L_k$ in Figure \ref{FIG::combined_plot_2} (B). There exists a strong negative correlation between attack effectiveness as measured by $\Delta A(L_k)$ and the attacked agent's signal strength. This is intuitive. If an agent has received a strong private signal in support of some value of $\theta$, a larger bias will be required to convince him of the contrary. Thus, if an attacker can target the biased signal to agents with weak private signals, the attack will be more effective.

Lastly, we consider the case of clustered vs.\ rewired/balanced networks and spread out vs.\ focused attacks. This analysis is done evaluating the baseline model under the biased signal attack in the clustered and balanced networks. A number of results are worth noting. As can be seen in Figure \ref{FIG::combined_plot_2} (C), for each scenario (baseline/no attack, focus and spread attacks) accuracy in the balanced network exceeds accuracy in the clustered network. This suggests that information aggregation is more effective in the balanced network. This is intuitive as the balanced networks has a shorter maximum path length between any two nodes thereby allowing information to ``propagate'' faster between nodes. In Figure \ref{FIG::combined_plot_2} (C), we can also see that variation in the accuracy between training runs with different seeds (each run corresponds to one marker for a particular scenario) is smaller for the balanced network. This suggests that in the balanced network learning good policies is an easier task than in the clustered network. We can also see from Figure \ref{FIG::combined_plot_2} (C) that spread attacks are more effective both in the clustered and balanced networks.

\subsection{Adapting to Attacks} 
So far we have evaluated models trained in the absence of an attack. However, one can expect that users of social networks will adapt over time to the presence of fake news. We therefore investigate to what extent citizens can learn to mitigate the effect of an attacker. For this purpose we train citizens in the presence of a biased signal attacker for the Barabasi-Albert network. We then evaluate the accuracy in the absence of an attack and in the presence of the attack under which they were trained. We summarize our results in Table \ref{TAB::Performance}.

\begin{table}[htbp]
\caption{Test accuracy of agents in the final time step ($t=T=10$) for different training and testing scenarios. All scenarios were run with the Barabasi-Albert graph. At test time without attack all agents receive unbiased signals and no agent is attacked. At test time with attack agents are exposed to the attack scenario they were trained under (except in the baseline case where attack is biased signal with $\beta=3$).}
\label{TAB::Performance}
\begin{center}
\begin{tabular}{lll}
  &\multicolumn{2}{c}{\bf Testing accuracy ($t=T$)}
\\ \cline{2-3} \\
\multicolumn{1}{c}{\bf Training scenario} & Without attack & With Attack \\
\\ \hline \\
Baseline (train: no attack, test: hand-tuned $\beta=3$) & 0.926 & 0.784\\
Biased signal attack ($\beta =3$) & 0.882& 0.849 \\
% Random action attack & 0.909& 0.908 \\
% Agent takeover attack & 0.863 & 0.844\\
\end{tabular}
\end{center}
\end{table}

Let us first contrast the accuracy of the baseline model (trained without attack) under a biased signal attacker with the accuracy of a model trained under this attack scenario. When trained in the presence of an attacker the test accuracy under attack increases from $0.78$ to $0.84$ relative to the baseline. This can also be seen in Figure \ref{FIG::combined_plot} (B). We conclude that agents can indeed learn to adapt to the presence of fake news. However, this adaptation comes at a cost. When trained in the presence of an attacker, the test accuracy without attack decreases from $0.92$ to $0.88$ relative to the baseline. We conjecture that this is because agents learn to trust strong private signals less and are less likely to follow their neighbors' actions. To summarize, agents can learn to adapt to the presence of an attacker which reduces attack effectiveness. However, adaptation comes at a cost. If trained under attack and tested in the absence of attack, accuracy is lower than in the baseline which is trained without attack.

\section{Experimental Design and Procedures}\label{exp_design}
%
% =======================================================================================================

To see whether the results obtained from the reinforcement learning model tell us something about actual human behavior in networks, we conduct a human-subject experiment. In this experiment, we test comparative statics obtained with the reinforcement learning model.

The experiment was pre-registered before the data collection began. The pre-registration is available at \url{https://aspredicted.org/blind.php?x=rf557t}. Data collection and data analysis were entirely separated.

% =======================================================================================================
\subsection{Course of Events}
% =======================================================================================================

In the beginning of the experiment, participants read the instructions. Before they are able to proceed to the first round, they also have to answer a set of comprehension test questions correctly. The instructions and comprehension test questions are reproduced in Appendix~\ref{app:instructions} and~\ref{app:test_questions}.

Participants are randomly assigned to groups of eight. The eight participants of one group are connected on a network (the type of network that the participants are connected on depends on the treatment and will be shown in Section~\ref{sub:treatments} below). The group composition remains the same during the whole experiment, and participants only interact with participants of their own group (directly or indirectly).  

The experiment consists of six rounds. Each round consists of ten periods. At the beginning of each round, the true state of the world is randomly determined to be either $-1$ or $1$, each with probability one half.\footnote{We choose $-1$ and $1$ in the experiment as states of the world instead of $0$ and $1$, so that negative signals correspond to signals pointing to one state of the world and positive signals to the other state.} The state of the world remains the same throughout all periods of a round. 

The participants' task is to guess the state of the world correctly in each of the ten periods of this round (the state of the world is framed neutrally as a coin flip in the experimental instructions). Thus, in each period, they have to make a binary choice, guessing $-1$ or $1$.\footnote{In the first round, there is a time limit of $60$ seconds per decision, in the other rounds of $30$ seconds per decision. However, the time limit is not strict, after that time a message pops up asking participants to make their choice.} The true state of the world in a round is not revealed to participants until the end of the experiment. 
Participants can draw on two types of information to make their guesses. First, they can use their own private signal. This private signal is drawn once for a round and is communicated to participants before the round starts. The signal is drawn from a normal distribution with standard deviation one and is independent across participants and rounds (the mean of a participant's signal depends on the treatment; the average of this mean across all participants of a group is always negative if the true state of the world is $-1$ and positive if the state of the world is $+1$, so that the signals jointly are always informative). 
The second type of information participants can use are the past actions of all participants that they are connected to on the network. For instance, when making the guess for the fifth period of a round, a participant connected to two other participants on the network knows the guesses of these two other players from periods one to four (in addition to her own private signal). 

In each period, a participant who guessed the state of the world correctly receives $100$ points. At the end of the experiment, one period per round is randomly selected for payment (this is clearly communicated to participants in the instructions; only one period per round is paid out, so that participants cannot hedge). Points are exchanged into euros at an exchange rate of $40$ points to $1$ EUR. In addition to the payoffs from the tasks, each participant receives a show-up fee of five euros.

% =======================================================================================================
\subsection{Treatments}\label{sub:treatments}
% =======================================================================================================

The experiment makes use of a $2 \times 3 $-factorial between-subjects design. The two treatment dimensions are the type of network (rewired network or clustered network) and the type of fake-news attack represented by the existence and type of biased signals (no attack, spread-out attack, concentrated attack). 

\begin{figure}[htbp]
\begin{center}
\subfloat{\includegraphics[width=0.5\textwidth, trim= 0.5cm 0cm 0cm 0cm, clip=false]{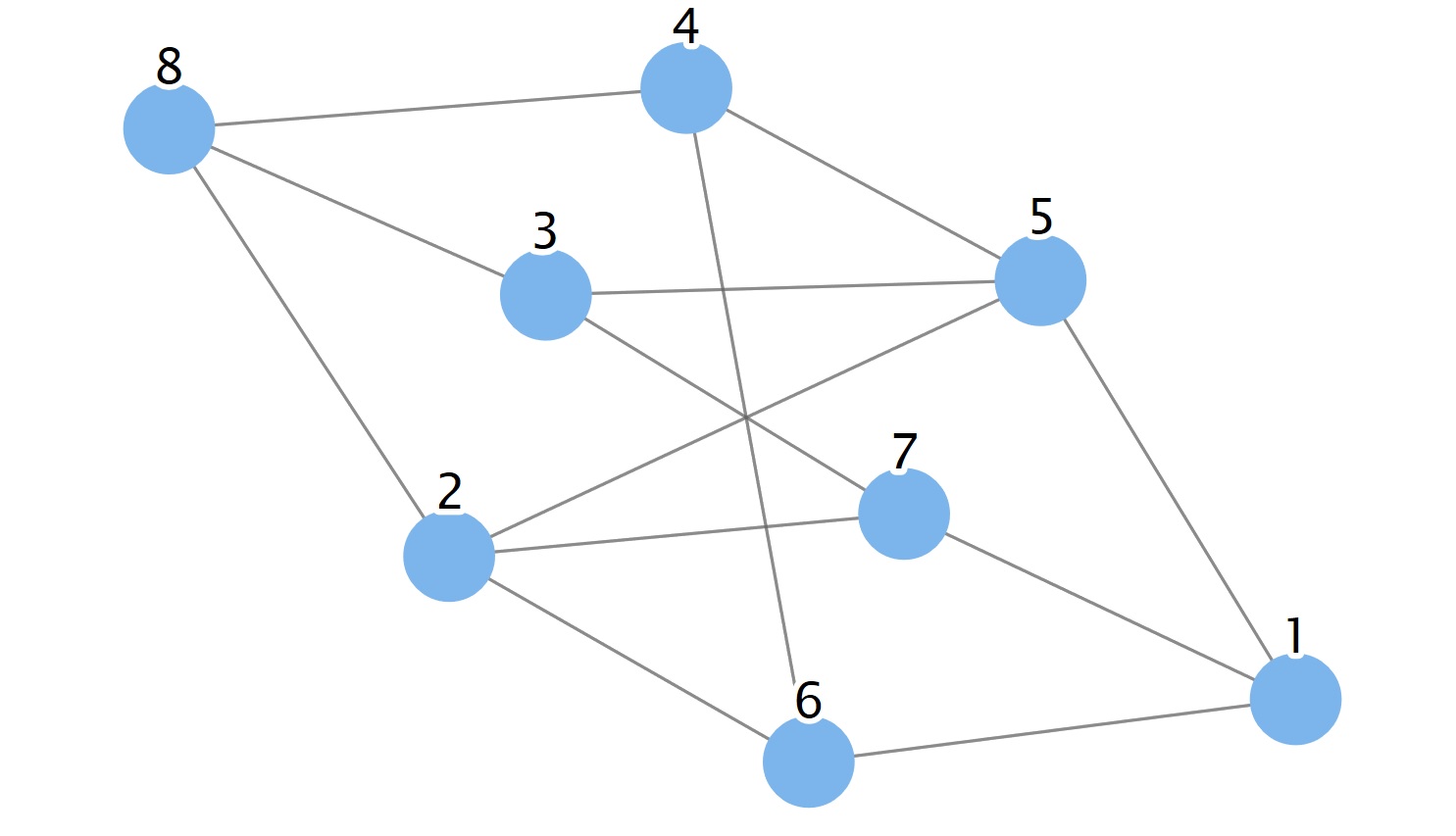}}
\subfloat{\includegraphics[width=0.4\textwidth, trim= 0cm 0cm 0cm 0.1cm, clip=true]{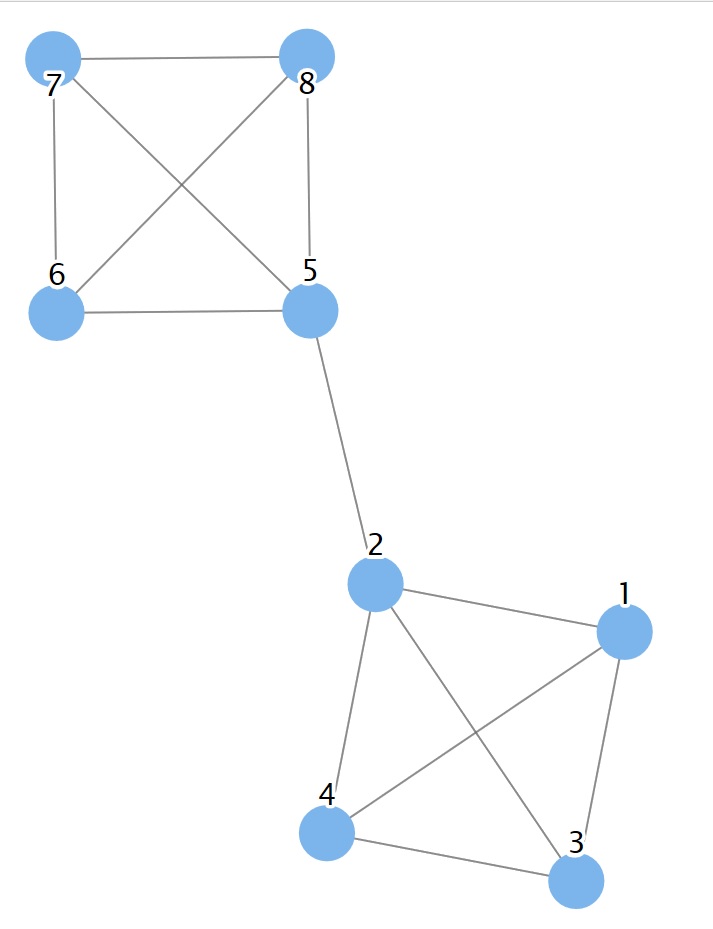}}
\end{center}
\caption{Rewired network (left) and clustered network (right), as used in the experiment.}\label{fig:networks_experiment}
\end{figure}

Figure~\ref{fig:networks_experiment} shows the networks used in the experiment. In the treatments without attack, all private signals are unbiased (thus, all signals are drawn from a normal distribution with the mean equal to the true state of the world). In the treatments with a spread-out attack, four members of the group receive an unbiased signal and four members receive a biased signal, where the bias is $1.75$ (that is, the four attacked members with the biased signal receive a draw from a normal distribution with mean $0.75$ if the state of the world is $-1$ and mean $-0.75$ if the state of the world is $1$). In the treatments with a concentrated attack, six group members receive an unbiased signal and two receive a biased signal with a bias of $3.5$ (thus, the two attacked members receive a signal from a normal distribution with mean $2.5$ if the state of the world is $-1$ and $-2.5$ if the state of the world is $1$).

The instructions contain the information that there may be biased signals---there is no deception in the experiment. However, we do not tell participants whether they are in a treatment with or without biased signals or how many biased signals of which strength there are. 

Table~\ref{tab:design_summary} summarizes the design. We number treatments consecutively: T1 is \textit{rewired\_no-attack}, T2 is \textit{rewired\_spread}, T3 is \textit{rewired\_concentrated}, T4 is \textit{clustered\_no-attack}, T5 is \textit{clustered\_spread}, and T6 is \textit{clustered\_concentrated}. 
We consider the treatments without attack, T1 and T4, control treatments. 
The number of observations corresponds to the number announced in the pre-registration.

\begin{small}
\begin{table}[htbp]
%\vspace{-6pt}
\begin{center}
\caption{Design Summary. This table shows the treatment abbreviations and the numbers of participants and networks (in parentheses) per treatment. }\label{tab:design_summary} 
%\vspace{6pt}
\begin{tabular}{lccc} \toprule %\toprule
& No attack & Spread-out attack & Concentrated attack \\ \midrule
Rewired network & T1, $56\ (7)$ & T2, $112\ (14)$ & T3, $112\ (14)$\\
Clustered network & T4, $56\ (7)$ & T5, $112\ (14)$ & T6, $112\ (14)$\\ \bottomrule
\end{tabular}
%\vspace{-8pt}
\end{center}
\end{table}
\end{small}

% =======================================================================================================
\subsection{Hypotheses}\label{sub:hypotheses}
% =======================================================================================================

Our outcome variable of interest is the accuracy of participants' guesses of the state of the world. The hypotheses are derived with the multi-agent reinforcement learning model (while the simulation results are shown in Section~\ref{sec:results} with different network sizes, the comparative statics---and thus the hypotheses---are identical when conducted with the networks used in the experiment). The hypotheses are as in the pre-registration and can be grouped into three parts (a to c below). The predictions by the model are our hypotheses, thus in statistical terms these are the alternative hypotheses. 

\begin{enumerate}[a)]
\item Accuracy is greater in the control treatments than in the treatments with fake news for the same network type: acc(T1) > acc(T2),  acc(T1) > acc(T3),  acc(T4) > acc(T5), and acc(T4) > acc(T6).
%\begin{enumerate}[label=a\arabic*.]
%\item  acc(T1) > acc(T2)
%\item  acc(T1) > acc(T3)
%\item  acc(T4) > acc(T5)
%\item  acc(T4) > acc(T6)
%\end{enumerate}
\item  Accuracy is greater in the treatment with concentrated attacks than in those with spread-out attacks for the same network type: acc(T3) > acc(T2) and  acc(T6) > acc(T5).
%\begin{enumerate}[label=b\arabic*.]
%\item  acc(T3) > acc(T2)
%\item  acc(T6) > acc(T5)
%\end{enumerate}
\item  Accuracy is greater in rewired networks than in clustered networks for the same type of attack: acc(T1) > acc(T4), acc(T2) > acc (T5) and acc(T3) > acc(T6).
%\begin{enumerate}[label=c\arabic*.]
%{\color{gray}\item  acc(T1) > acc(T4)}
%\item  acc(T2) > acc (T5)
%\item  acc(T3) > acc(T6)
%\end{enumerate}
\end{enumerate}

Note that for hypothesis c1, acc(T1) > acc(T4), our model indicates that the difference between the two accuracies is so small that we did not expect to find statistically significant results.\footnote{This is written in the pre-registration as follows: {\em ``These are the predictions/hypotheses from the model. Note, however, that we expect the treatment difference "c1. acc(T1) > acc(T4)" to be so small that we do not expect to observe statistically significant results (this is a reason for choosing a lower number of observations in the control treatments than in the other treatments...).''}}

The hypotheses vary in terms of how likely it is that they will be confirmed. The hypothesis set a) comparing the control treatments to the the treatments with fake news---that is, with biased signals---are likely to be confirmed in experiments. These hypotheses do not require a high degree of rationality by participants to hold.
Hypothesis sets b) and c) are less obvious and it is easy to come up with theories pointing towards the opposite. For instance, if those citizens with particularly strong signals are most likely to stick with their signals, while other citizens are more likely to react to others' actions, this could lead to concentrated attacks being more successful (i.e., leading to lower accuracy) than spread-out attacks, in contrast to our predictions. It would also be conceivable that that attacks are more successful in rewired networks than in clustered networks (in contrast to our predictions), as the biased signals could spread to a larger set of participants immediately instead of propagating first within a cluster.

% =======================================================================================================
\subsection{Procedures}\label{sub:procedures}
% =======================================================================================================

We report data from a total of $560$ participants. $56$ experimental subjects (seven independent networks of eight) participated in each of the two control treatments. $112$ subjects ($14$ independent networks) participated in each of the four fake-news treatments.\footnote{The treatments were balanced across the experimental sessions, so that in each session there were zero or one networks of each control treatment and one or two networks of each fake-news treatment (as fewer data points were collected from the control treatments).} The experiment was conducted at the LINEEX laboratory in Valencia. The experiment was carried out in Spanish (instructions and comprehension test questions in the appendix are the English versions; the translation to Spanish was carefully verified). After the experiment, participants filled out a short questionnaire asking for a few demographic variables. Approximately half of the participants were female. The average age of participants was about $21$ years. About half of them were students of economics, finance, or business. More than $90\%$ of participants were Spanish.

% =======================================================================================================
%
\section{Experimental Results}\label{exp_results}
%
% =======================================================================================================

The key outcome variable we use is the mean accuracy in a network; that is, the mean across participants on the network across all rounds and periods. As participants only interact with participants in the same network, these network-level variables are statistically independent. We rely on Wilcoxon rank-sum tests for hypothesis testing, using one-sided tests because of the directedness of our pre-registered hypotheses. 
Robustness checks making use of regressions can be found in Appendix~\ref{subapp:additional_data}. 
The statistical analysis generally follows the description in the pre-registration.\footnote{The pre-registration mentions the mean accuracy across all periods and rounds as first outcome measure. In addition, the pre-registration mentions the mean across all periods of the last round only as a second outcome measure, a measure that is more noisy than the first measure and that we have therefore discarded.} 

\begin{figure}[htbp]
\begin{center}
\includegraphics[width=\textwidth, trim= 0.0cm 0.75cm 0cm 0cm, clip=false]{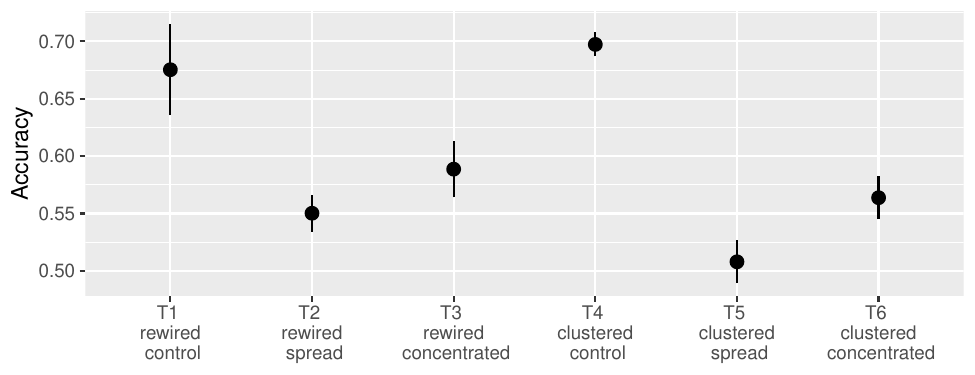}
\end{center}
\caption{Mean accuracy in the treatments of the experiment.}\label{fig:accuracy_experiment}
\end{figure}

Figure~\ref{fig:accuracy_experiment} shows the experimental results, depicting the mean accuracy per treatment. Table~\ref{tab:p-values} shows the hypotheses discussed above (i.e., the relationships predicted by the reinforcement learning model), together with the relationship observed in the data and $p$-values of Wilcoxon rank-sum tests. 

\begin{table}[htbp]
%\vspace{-6pt}
\begin{center}
\caption{Comparative statics in predictions and data and $p$-values. This table shows the comparative statics of the predictions, i.e. the hypotheses, jointly with the directions of treatment effects, jointly with the $p$-values of the one-sided Wilcoxon rank-sum tests.\ $^*$ For this hypothesis, no significant differences were expected ex ante, as discussed in Section~\ref{sub:hypotheses} and the pre-registration.}
\label{tab:p-values} 
%\vspace{6pt}
\begin{tabular}{lccc} \toprule %\toprule
& Hypothesis & Data & $p$-value \\ \midrule
\multicolumn{4}{l}{Control (left) vs.\ fake news (right)} \\ \midrule
a1.  acc(T1) vs.\ acc(T2) & > & > & $0.003$ \\
a2. acc(T1) vs.\ acc(T3) & > & > & $0.078$ \\
a3.  acc(T4) vs.\ acc(T5) & >  & > & $0.000$ \\
a4. acc(T4) vs.\ acc(T6) & > & > & $0.001$ \\ \midrule
\multicolumn{4}{l}{Concentrated (left) vs.\ spread (right)} \\ \midrule
b1.  acc(T3) vs.\ acc(T2) & >  & > & $0.135$ \\
b2.  acc(T6) vs.\ acc(T5) & > & > & $0.020$ \\ \midrule
\multicolumn{4}{l}{Rewired (left) vs.\ clustered (right)} \\ \midrule
{\color{black}c1.  acc(T1) vs.\ acc(T4)}$^*$ & {\color{black}>}  & {\color{black}<} & {\color{black}$0.896$} \\
c2.  acc(T2) vs.\ acc (T5) & >  & > & $0.047$ \\
c3.  acc(T3) vs.\ acc(T6) & >  & > & $0.275$ \\ \bottomrule
\end{tabular}
%\vspace{-8pt}
\end{center}
\end{table}

The directions of treatment effects from the experimental data are always as predicted, with the exception of the comparison of the two control treatments where the results are very close together and statistically highly insignificant. This is the hypothesis for which we already mentioned in the pre-registration that we do not expect to find support for it, as also discussed in Section~\ref{sub:hypotheses}. Note that for this comparison, there is only data from $14$ groups in total, less than for the other comparisons. Considering the three blocks of hypotheses, there is strong evidence that accuracy is lower in the fake-news treatments than in the control treatments. Three of the four tests are strongly significant while the fourth is marginally significant. 
There is also support that accuracy is lower when the fake-news attack is spread-out across more different participants with lower intensity than when it is concentrated on fewer participants. This comparison is statistically significant at the $5\%$-level for the comparison when the network is clustered, while the $p$-value is slightly above the level for marginal significance in the rewired-network comparison. 
There is also some evidence that accuracy is lower in clustered networks than in rewired networks in the presence of a fake-news attack. When the fake-news attack is more effective (i.e., when it is spread-out), this difference is statistically significant. When this fake-news attack is less effective (i.e., when it is concentrated), the direction of the treatment effect points in the same direction, while the treatment difference is statistically insignificant. 

Overall, not all hypothesized treatment differences are statistically significant, but basically all treatment directions are as hypothesized, and most of the treatment differences are statistically significant (at least at the 10\% level). In sum, our experimental results support the comparative statics obtained with the reinforcement learning model in Section \ref{sec:results}.

% =======================================================================================================
%
\section{Concluding Remarks}\label{sec:conclusion}
%
% =======================================================================================================
The deliberate manipulation of the public's perception of facts via fake news, in particular on social networks, has become a growing concern for policy makers and technology companies alike. Despite its importance, little is understood about the spread of fake news on social networks. This is in large part due to the technical difficulties involved in studying the complex decision making on social networks. We develop a framework for the study of fake news in social networks that is theoretically grounded yet computationally tractable and flexible. We achieve this by first extending a standard game of information aggregation to accommodate fake news and then applying state-of-the-art deep multi-agent reinforcement learning to solve the game. We use predictions of the model to form hypotheses that we can test in a controlled laboratory experiment with human subjects. In our experiment, we find evidence that our model of human behaviour is indeed useful to understand the spread of fake news in social networks.

Our findings suggest a number of interventions that can contribute to making fake news attacks in social networks less effective. First, social media companies could adjust their algorithms suggesting users which other users to connect to (such algorithms exist for many social media platforms, including Facebook, LinkedIn and Twitter/X). These algorithms could give preference to connections that are in different clusters of the network, thereby making the network more balanced. 
Second, keeping some information on social networks private, such as the number of connections, can make it harder for attackers to condition their attacks on the network position. % and general informedness (as the real world analogue of signal strength). This should reduce attack effectiveness. 
In addition, simply making agents aware of the potential presence of fake news will already contribute to their adaption and mitigate the effect of disinformation.
%However, this adaptation is likely to be accompanied by evolving attack strategies, so any adaptation will not be lead to permanent mitigation. In addition, the adaptation is likely to harm information aggregation in the absence of fake news. 

%Third, encouraging well balanced social networks can improve information aggregation in general and make fake news attacks less effective in particular. 

%Our current approach has two main shortcomings. First, it does not scale well to larger populations of agents since it requires the entire social network to be trained simultaneously. One way forward is to train agents in sub-graphs feeding them an embedding of the graph, see for example \cite{gilmer2017neural}, and then composing larger graphs at test time. The second shortcoming is the use of IQL in the agent takeover attack scenario. While the lack of convergence leads to interesting qualitative insights, MARL methods that encourage convergence of policies would be desirable here, at least as benchmarks. One possible way to achieve this would be to apply stable opponent shaping to citizen and attacker, see \cite{letcher2018stable}.  This provides interesting avenues for future research.

% =======================================================================================================
%
\bibliographystyle{apalike}
\bibliography{references}
%
% =======================================================================================================

% =======================================================================================================
%
% APPENDIX
%
% =======================================================================================================
\pagebreak

\appendix
\numberwithin{equation}{section}
\numberwithin{figure}{section}
\numberwithin{table}{section}

\pagenumbering{arabic}% resets `page` counter to 1
\renewcommand*{\thepage}{A\arabic{page}}

\begin{center}
\huge{Online Appendix to ''Fake News in Social Networks''}
\end{center}
\setcounter{footnote}{0}

\begin{center}
Christoph Aymanns \qquad Jakob Foerster \qquad Co-Pierre Georg \qquad Matthias Weber
\end{center}

%\vspace{1cm}

This online appendix contains material in addition to the main text. Section~\ref{app:instructions} contains the experimental instructions. Section~\ref{app:test_questions} reproduces the comprehension test questions used in the experiment. Section~\ref{subapp:additional_data} contains additional statistical analyses of the experimental data as robustness check.

% subsection implementation_details (end)

% subsection multi_agent_reinforcement_learning (end)

% \subsection{Additional figures} % (fold)
% \label{sub:additional_figures}

% \begin{figure}[h]
% \begin{center}
% %\framebox[4.0in]{$\;$}
% \includegraphics[width=0.45\textwidth]{figs/decline_in_acc_by_attacker_neighbor_signal_st.pdf}
% \end{center}
% \caption{Effectiveness of attack when conditioning on the average signal strength of neighbors of the attacked agent. All agents are trained in the absence of any attacker, i.e. with unbiased signals only, on a Barabasi-Albert graph with $|N|=10$. Decline in accuracy is measured as the difference of the accuracy under attack (a single agent receives a biased signal ($\beta=3$)) to the baseline without attack. Conditioning on the average signal strength of the neighbors of the attacker: The signal strength is the absolute value of the log-likelihood ratio of $\theta$ given the signal. For each agent, the average signal strength of her neighbors is computed. For the analysis signal strengths were binned into deciles. Plot in (b) shows decline in accuracy averaged over signal strength decile bins on the y-axis and the signal strength decile mid point on the x-axis.}\label{FIG::neighbor_signal}
% \end{figure}

% =======================================================================================================
%
\section{Experimental Instructions}\label{app:instructions}
%
% =======================================================================================================
We first reproduce the complete instructions of the treatments with rewired networks, T1-T3, in Section~\ref{subapp:instructions_rewired}. We then describe where the instructions of the treatments with clustered networks, T4-T6, differ from the reproduced instructions. These differences are described in Section~\ref{subapp:instructions_differences}.

% =======================================================================================================
\subsection{Instructions in the Treatments with Rewired Networks}\label{subapp:instructions_rewired}
% =======================================================================================================

\textbf{\large{Instructions}}

Please read these instructions carefully as they explain how you earn money from the decisions that you make. You will be paid privately at the end, after all participants have finished the experiment.

\textbf{During the experiment you are not allowed to use the internet, your mobile phone, tablet, or notebook. You are also not allowed to communicate with other participants.}

If you have a question at any time, please raise your hand and someone will come to your desk to answer your question in private.

\textbf{The experiment consists of 6 rounds. Each round consists of 10 periods.} Your decision in each period and round may affect your final payoff. \textbf{At the end, only your earnings from one randomly selected period per round will be paid out to you.} This means that for each round, one period will be randomly selected, and the payment from the experiment will be the sum of these 6 randomly selected periods (one per round). The points from the selected period will be exchanged into EUR at the exchange rate 40 points = 1 EUR. In addition, you will receive 5 EUR for your participation.

You have been randomly assigned to a \textbf{group of 8 participants/players} (including you). Your group will not change during the experiment. You will not know the identity of any other group member nor will they know your identity even after the experiment is over.

The following describes what you will be doing in \textbf{each} of the 6 rounds. \\

\textbf{\large{General information}}

In the experiment, all members of one group are connected in a network. Below, you can see this network. The network will be the same throughout the whole experiment. Within each group, players are assigned numbers from 1 to 8. As shown in the graph below, the numbers determine the positions in the network. Each player will get to know their player number during the experiment. This player number and the position in the network will remain the same throughout the experiment. The lines in the graph indicate which players are connected. A connection between two players means that the players can observe each other's past decisions in the current round (what decisions players can make will be explained later).

\begin{figure}[htbp]
\begin{center}
%\framebox[4.0in]{$\;$}
\includegraphics[width=0.7\textwidth]{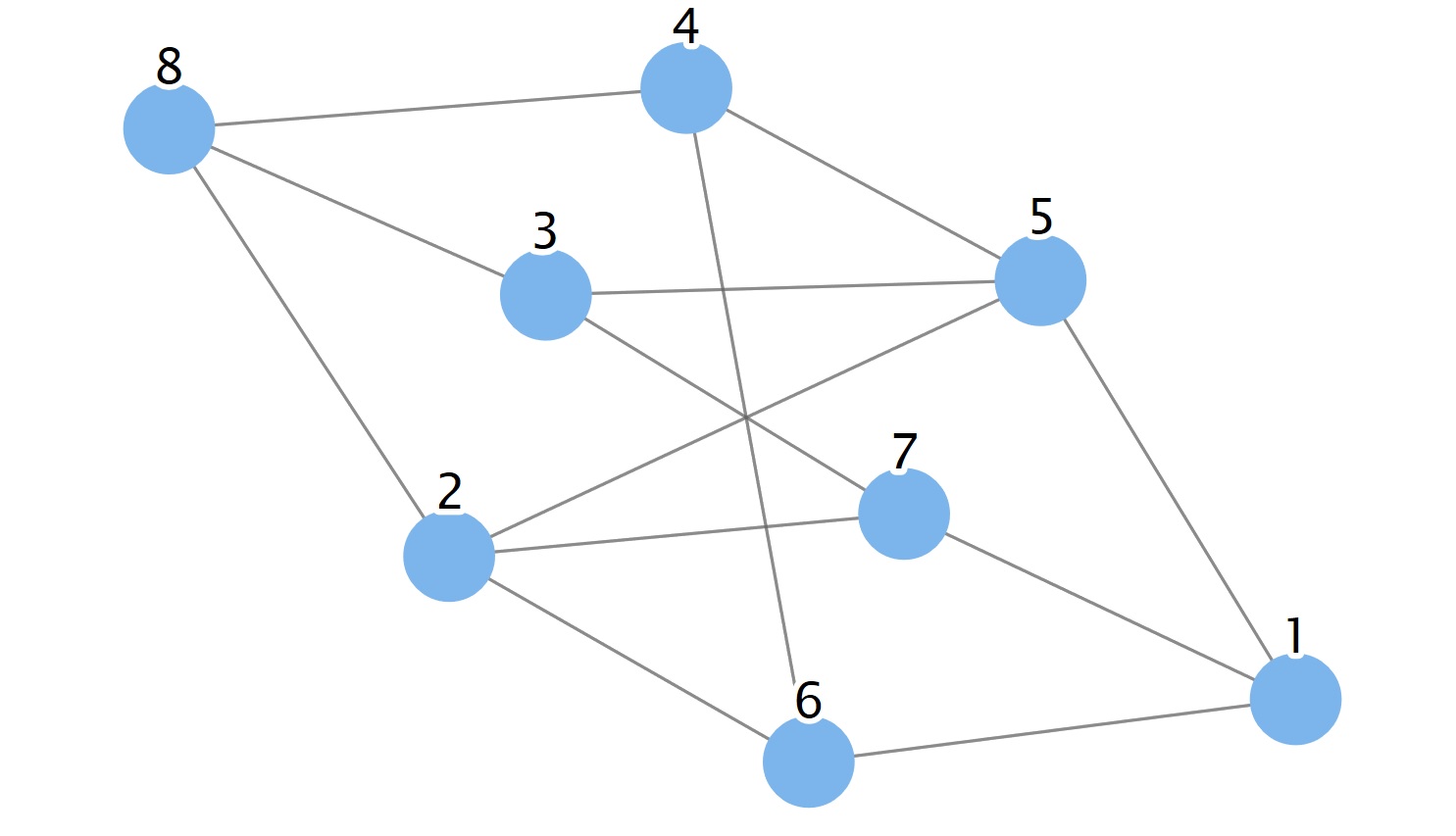}
\end{center}
\caption{Rewired network (not labeled in the experimental instructions).}\label{FIG:network_rewired}
\end{figure}

\begin{center}[Figure~\ref{FIG:network_rewired} appears here in the experimental instructions.]\end{center}

In this network, player 7 can for example observe the past decisions of players 1, 2, and 3 but not the decisions of other players. As another example, player 2 can observe the past decisions of players 5, 6, 7, and 8. \\

\textbf{\large{Setting, decisions, payments}}

At the beginning of each round, we flip a fair coin (to be precise, the computer's random number generator ``flips the coin''). On one side of the coin, the number 1 is written. On the other side of the coin, the number -1 is written. That is, 1 and -1 appear with equal probability.
Each player's task is to correctly guess the outcome of the coin flip (later, we will explain which hints/information you will receive that may be relevant for your guesses).

In each period (of each round), you need to state whether your guess of the outcome of the coin flip is -1 or 1. If your guess of the number in a period was correct, you receive 100 points, if the guess was incorrect you receive nothing for that round (remember that only one period of each round will later be randomly selected for payment). Note that the outcome of the coin flip for any round will not be revealed to you until the very end of the experiment when the payments are determined.

In the first round, you have up to 60 seconds for each decision. In the other rounds, you have up to 30 seconds. When the time elapses, a notification pops up (you will still be able to make a decision after this notification, but we ask you to respond within the given time in order not to prolong the experiment). \\

\textbf{\large{Signals and information}}

At the beginning of each round (after the coin has been flipped), each player receives a signal about the outcome of the coin flip. This signal has a random component in it.

Two types of signals exist: unbiased signals and biased signals. The mean (that is, the long-term average) of the unbiased signal is equal to the outcome of the coin flip. The mean of the biased signal is not equal to the outcome of the coin flip. Both unbiased signals and biased signals follow a so-called normal distribution with standard deviation 1. The signals are drawn independently for all players (this means that the random disturbance added to the mean in one player's signal does not influence the random disturbance added to the mean of another player's signal).

At least half of the players of a network receive an unbiased signal. That is, at least four players in a network receive a signal that is on average correct. You do not know whether your signal is biased or unbiased. Note that the average of the mean of the 8 signals in one group is always closer to the outcome of the coin flip than to the number that did not come up in the coin flip. In simple terms, this means that you can always expect that all of the group's signals considered jointly are informative about the outcome of the coin flip.

The next graph shows how likely it is that such a signal falls into a certain range. In this example, the outcome of the coin flip (that players cannot observe) is 1. The probability that the random signal of any player who receives an unbiased signal falls into a certain range of numbers corresponds to the area under the curve. For example, the gray shaded area shows how likely it is that the outcome of this random signal falls into the range from 1.5 to 2.5.

\begin{figure}[htbp]
\begin{center}
%\framebox[4.0in]{$\;$}
\includegraphics[width=0.7\textwidth]{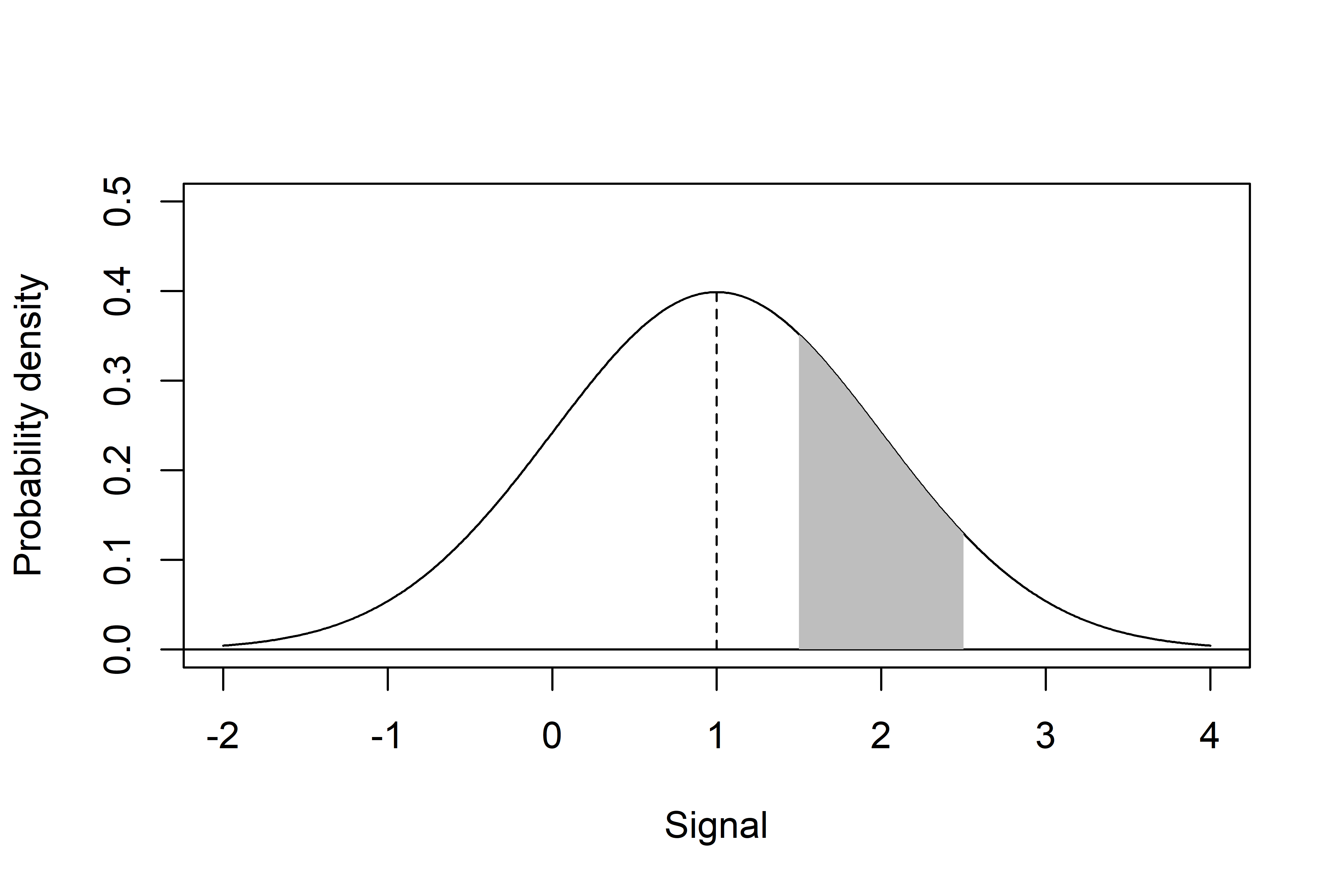}
\end{center}
\caption{Density of a normal distribution (not labeled in the experimental instructions).}\label{FIG:normal_dist}
\end{figure}

\begin{center}[Figure~\ref{FIG:normal_dist} appears here in the experimental instructions.]\end{center}

On your screen, you will always see your signal for the ongoing round, your past decisions in that round, and the past decisions of the players connected to you. The following overview shows the course of events in \textbf{each of the 6 rounds}:

\begin{itemize}
\item The outcome of the coin flip is determined (but not revealed to any of the players)
\item Each player receives a signal.
\item Each player enters his/her guess (-1 or 1) for period 1.
\item Each player receives the information what players connected to him/her guessed previously. Each player enters his/her guess (-1 or 1) for period 2.
\item The previous step is repeated until the guess for period 10 has been entered. Then the round terminates.
\end{itemize}

% =======================================================================================================
\subsection{Differences in Instructions in the Treatments with Rewired Networks}\label{subapp:instructions_differences}
% =======================================================================================================

The instructions of the treatments with clustered networks, T4-T6, are identical to the instructions reproduced in Section~\ref{subapp:instructions_rewired}, except for the graph of the network and its description. The graph from treatments T4-T6 is shown in Figure~\ref{FIG:network_clustered}.

\begin{figure}[htbp]
\begin{center}
%\framebox[4.0in]{$\;$}
\includegraphics[width=0.5\textwidth, trim= 0 0 0 0.1cm, clip = true]{figs/network_clustered.jpg}
\end{center}
\caption{Clustered network (not labeled in the experimental instructions).}\label{FIG:network_clustered}
\end{figure}

The description of the graph in the instructions of the clustered treatments is as follows: ``In this network, player 7 can for example observe the past decisions of players 5, 6, and 8 but not the decisions of other players. As another example, player 2 can observe the past decisions of players 1, 3, 4, and 5.''

% =======================================================================================================
%
\section{Comprehension Test Questions}\label{app:test_questions}
%
% =======================================================================================================

Below, we reproduce the comprehension test questions that all participants had to answer correctly before starting the experiments. The questions were the same in all treatments. We indicate the correct answer with a check mark. 

Choose the correct option.

1. At the end of the experiment, one period of each round will be randomly selected and your payment then depends on...
\begin{enumerate}[a)]
\item your decisions in these 6 selected periods and the outcomes of the coin flip (to be precise, your payment only depends on whether your guesses in these 6 selected periods were correct). \checkmark
\item your decisions and others' decision, not on the outcomes of the coin flips.
\item luck. Your payment does not depend in any way on your decisions in the experiment.
\end{enumerate}

2. 
\begin{enumerate}[a)]
\item You can observe the past decisions of all other players in your network.
\item You can observe the past decisions of the players that you are connected to on the network, but you do not know who can observe your past decisions.
\item You can observe the past decisions of the players that you are connected to on the network, and exactly those players can observe your past decisions. \checkmark
\end{enumerate}

3.
\begin{enumerate}[a)]
\item The outcome of the coin flip is always the same for all players in the network. However, it can be different from period to period and from round to round.
\item The outcome of the coin flip is always the same for all players in the network. It is also the same in all periods of a round. However, it can be different from round to round. \checkmark
\item The outcome of the coin flip may be different for different players in the network. However, it is the same for each player across periods and rounds.
\end{enumerate}

4. The signal about the outcome of the coin flip is...
\begin{enumerate}[a)]
\item on average correct for at least four players in the network (where "on average" refers to the theoretical mean of the probability distribution). This may be any number from four to eight throughout the whole experiment. The number may also vary from round to round, as nothing is written about this in the instructions. \checkmark
\item on average correct for at least four players in the network (where "on average" refers to the theoretical mean of the probability distribution). This number is necessarily different from round to round.
\item never less than -1 and never greater than 1.
\end{enumerate}

5. If a player receives a biased signal,...
\begin{enumerate}[a)]
\item the mean of this signal is -1 if the outcome of the coin flip was 1 and 1 if the outcome of the coin flip was -1.
\item the mean of this signal is unknown to you, as the instructions do not provide an exact number for it. \checkmark
\item at least four other players' signals are also biased in this round.
\end{enumerate}

6. The instructions contain the following sentence: "Note that the average of the mean of the 8 signals in one group is always closer to the outcome of the coin flip than to the number that did not come up in the coin flip." What does this mean in simple terms?
\begin{enumerate}[a)]
\item You can always expect that the average of all signals in a group equals either 1 or -1. Whether you can expect this average to equal 1 or -1 depends on the outcome of the coin flip.
\item You can always expect that all of the group's signals considered jointly are informative about the outcome of the coin flip. \checkmark
\item All 8 signals are necessarily between -1 and 1.
\end{enumerate}

% =======================================================================================================
%
\section{Robustness Check: Regression Analyses of the Experimental Data}\label{subapp:additional_data}
%
% =======================================================================================================

As a robustness check to the statistical analysis in Section~\ref{exp_results}, we present here the results from regression analyses. In a specification without demographic control variables, we estimate the following equation with OLS:

\begin{align*}
ACCURACY_i &= \beta_0 + \beta_1 CLUST_i + \beta_2 FAKE\_NEWS_i + \beta_3 SPREAD_i + \beta_4 CLUST_i * FAKE\_NEWS_i \\
&+ \beta_5 CLUST_i * SPREAD_i  + \eps_i \numberthis \label{regression_spec}
\end{align*}

The outcome variable $ACCURACY$ is the average accuracy in a network (across all periods and rounds, as in the analysis in Section~\ref{exp_results}). There is thus only one observation per network and observations are statistically independent. $CLUST$ is a binary variable taking values $0$ (in the treatments with the rewired network) or $1$ (in the treatments with the clustered network). $FAKE\_NEWS$ is $0$ in the control treatments, otherwise $1$. $SPREAD$ equals $1$ in the two treatments with spread-out attacks, otherwise $0$. 

In a specification including demographic control variables, the right side of Equation~\eqref{regression_spec} additionally includes the term $+ \beta_6 GENDER + \beta_7 AGE + \beta_8 STUDIES$, where $GENDER$ represents the fraction of females in a network, $AGE$ the average age, and $STUDIES$ the fraction of students of economics, finance, or business.

We report the coefficient estimates of these two regressions in Table~\ref{tab:regression_coefficients}. Table~\ref{tab:tests_regressions} is similar to Table~\ref{tab:p-values} in the main text. In Table~\ref{tab:tests_regressions}, we reiterate our hypotheses and show to which sign of the sums of regression coefficients these hypotheses lead in the regressions (in the table, we use the variable names rather than the estimated coefficients, because this makes clearer what is meant; e.g., we write $FAKE\_NEWS+SPREAD$ instead of $\beta_2 + \beta_3$). The column ``Prediction'' states what needs to hold for the sum of the coefficients if the prediction/hypothesis holds. The following column contains the relation in the regressions (one column for both regression specifications, because it turns out that these observed directions of treatment effects are always identical for both specifications). Showing the statistical significance of the obtained relationships, the last two columns contain the $p$-values of (one-sided) tests of the relationships, separately for the specifications without and with demographic control variables.

%\begin{small} 
\begin{table}[htbp]
\begin{center}
%\onehalfspacing
\caption{Regression coefficient estimates. This table reports the coefficient estimates from OLS regressions without (left) and with (right) demographic control variables. The outcome variable is the average accuracy in a network across periods and rounds.}\label{tab:regression_coefficients}
%\vspace{-6pt}
\begin{tabular}{lcc}
\toprule
& w/o dem.\ controls & w dem.\ controls  \\
\midrule
INTERCEPT & $ 0.675$ & $0.579$ \\
CLUST & $0.022$ & $0.017$  \\ 
FAKE\_NEWS & $-0.087 $ & $-0.094$ \\ 
SPREAD & $-0.038$ &   $-0.034$  \\ 
CLUST * FAKE\_NEWS & $-0.047$ & $ -0.041 $ \\ 
CLUST * SPREAD & $-0.017$ & $-0.012$ \\ %\multicolumn{2}{l}{No.\ female} 
GENDER & & $ -0.078$ \\ 
AGE & & $0.007$ \\ 
STUDIES & & $0.004$ \\ \midrule
$R^2$& $0.413$ &  $0.442$   \\ 
$N$ & $70$ & $70$ \\ 
\bottomrule
\end{tabular}
\end{center}
\end{table}
%\end{small}

\begin{small}
\begin{sidewaystable}[htbp]
%\vspace{-6pt}
\begin{center}
\caption{Comparative statics in predictions (hypotheses) and tests on regressions coefficients}
\label{tab:tests_regressions} 
%\vspace{6pt}
\begin{tabular}{llcccc} \toprule %\toprule
Hypothesis & Regression coefficients & Prediction & Coeff.\ w/o \& w/ & $p$-value w/o & $p$-value w/ \\ 
 &  &  & dem.\ controls & dem.\ controls & dem.\ controls\\ \midrule
\multicolumn{6}{l}{Control (left) vs.\ fake news (right)} \\ \midrule
a1.  acc(T1) > acc(T2) & FAKE\_NEWS + SPREAD& $<0$ & $<0$   & $0.000$ & $0.000$\\
a2. acc(T1) > acc(T3) & FAKE\_NEWS & $<0$ & $<0$ & $0.007$ & $0.004 $\\
a3.  acc(T4) > acc(T5) & FAKE\_NEWS + SPREAD & $<0$  & <0 & $0.000$ & $0.000 $ \\
 & + CLUST * FAKE\_NEWS  & & & & \\
  & + CLUST * SPREAD & & & & \\
a4. acc(T4) > acc(T6) & FAKE\_NEWS + CLUST * FAKE\_NEWS & $<0$ & $<0$ & $0.000$ & $0.000 $ \\ \midrule
\multicolumn{6}{l}{Concentrated (left) vs.\ spread (right)} \\ \midrule
b1.  acc(T3) > acc(T2) & SPREAD & $<0$  & $<0$ & $0.088$ & $0.115 $ \\
b2.  acc(T6) > acc(T5) & SPREAD + CLUST * SPREAD & $<0$ & $<0$ & $0.026$ & $0.056 $ \\ \midrule
\multicolumn{6}{l}{Rewired (left) vs.\ clustered (right)} \\ \midrule
%{\color{gray}c1.  acc(T1) > acc(T4)}$^*$ & CLUST & {\color{gray}$<0$}  & 
%{\color{gray}$>0$} & {\color{gray}$0.709$} & {\color{gray}$0.666$} \\
c1.  acc(T1) > acc(T4)$^*$ & CLUST & $<0$  & $>0$ & $0.709$ & $0.666$ \\
c2.  acc(T2) > acc (T5) & CLUST + CLUST * FAKE\_NEWS  & $<0$  & $<0$  & $0.069$ & $0.108 $\\
 & + CLUST * SPREAD & & & & \\
c3.  acc(T3) > acc(T6) & CLUST + CLUST * FAKE\_NEWS & $<0$  & $<0$ & $0.190$ & $0.202 $\\ \bottomrule
\multicolumn{6}{l}{\begin{minipage}{22cm}~\\ \footnotesize This table shows the predictions (i.e., the hypotheses), formulated in terms of the regression specification, jointly with the results of these sums of coefficient estimates in the data and with (one-sided) $p$-values from the regressions (both without and with demographic control variables).\ $^*$ For this hypothesis, no significant differences were expected ex ante, as discussed in Section~\ref{sub:hypotheses} and the pre-registration. \end{minipage}}
\end{tabular}
\vspace{-8pt}
\end{center}
\end{sidewaystable}
\end{small}

The tests concerning the first four hypotheses are strongly significant, the level of significance of the tests of the other hypotheses is a bit lower and depends on the chosen regressions specification. By and large, the regression analysis confirms the findings discussed in the main text. 

\end{document}